\documentclass{article}

\makeatletter\newcommand*{\IfPackageLoaded}{\@ifpackageloaded}\makeatother

\usepackage{bigstrut}
\usepackage{booktabs}
\usepackage{breakcites}
\usepackage{delimset}
\usepackage{float}
\IfPackageLoaded{hyperref}{\hypersetup{colorlinks=true, bookmarks=true, citecolor=Maroon, linkcolor=Maroon, urlcolor=blue}}{\usepackage[colorlinks=true, bookmarks=true, citecolor=Maroon, linkcolor=Maroon, urlcolor=blue]{hyperref}}
\usepackage[hybrid]{markdown}
\usepackage{microtype}
\usepackage{mleftright}\mleftright
\usepackage{multirow}
\usepackage{nicefrac}
\usepackage{subcaption}
\usepackage[normalem]{ulem}
\usepackage{wrapfig}
\usepackage[svgnames]{xcolor}
\usepackage{zref}

\IfPackageLoaded{enumitem}{}{\usepackage[inline, shortlabels]{enumitem}}

\newtoggle{include-notes}
\newtoggle{error-notes}
\toggletrue{include-notes}
\toggletrue{error-notes}

\newcommand*{\B}[1]{\ifmmode\bm{#1}\else\textbf{#1}\fi}
\newcommand*{\C}[1]{\mathcal{#1}}

\newcommand*{\R}[1]{\mathrm{#1}}

\newcommand*{\Z}[1]{\mathds{#1}}

\DeclareMathOperator{\D}{\Z{D}}

\DeclareMathOperator{\E}{\Z{E}}

\newcommand*{\ccell}[1]{\omit\hfil$\displaystyle#1$\hfil}

\newcommand*{\eqn}[1]{\begin{align}#1\end{align}}

\newcommand*{\nn}{\nonumber}

\makeatletter
\def\fn{\gdef\@thefnmark{}\@footnotetext}
\makeatother

\makeatletter
\zref@newprop{mzheight}[0pt]{\the\ht\z@}
\zref@newprop{mzdepth}[0pt]{\the\dp\z@}
\newcount\c@@mz
\newcommand*{\the@mz}{mz\the\c@@mz}
\newcommand*{\@mz@list}{}    
\let\@mz@do\relax
\newcommand*{\mzreset}{%
  \begingroup
    \def\@mz@do##1{%
      \global\expandafter\let\csname mz@##1\endcsname\relax
    }%
    \@mz@list
    \global\let\@mz@list\@empty
  \endgroup
}
\newcommand*{\mzleft}[3]{%
  \@ifundefined{mz@#1}{%
    \global\advance\c@@mz\@ne
    \expandafter\xdef\csname mz@#1\endcsname{\the@mz}%
    \xdef\@mz@list{\@mz@list\@mz@do{#1}}%
  }{}%
  \expandafter\let\expandafter\@mz\csname mz@#1\endcsname
  \mleft#2%
  \expandafter\mathpalette\expandafter{%
    \expandafter\@mzleft\expandafter{\@mz}%
  }{#3}%
  \mright.\kern-\nulldelimiterspace
}
\newcommand*{\mzright}[3]{%
  \kern-\nulldelimiterspace
  \@ifundefined{mz@#1}{%
    \@latex@warning{Missing \string\mzleft{#1}}%
    \mleft.#2\mright#3%
  }{%
    \expandafter\let\expandafter\@mz\csname mz@#1\endcsname
    \mleft.%
    \expandafter\mathpalette\expandafter{%
      \expandafter\@mzright\expandafter{\@mz}%
    }{#2}%
    \mright#3%
  }%
}   
\newcommand*{\@mzleft}{%
  \@mzleftright lr%
}
\newcommand*{\@mzright}{%
  \@mzleftright rl%
}
\newcommand*{\@mzleftright}[5]{%
  \sbox0{$\m@th#4{}#5{}$}%
  \ifmeasuring@
  \else
    \begingroup
      \let\@auxout\@mainaux
      \zref@labelbyprops{#3#1}{mzheight,mzdepth}%
    \endgroup
  \fi
  \zifrefundefined{\@mz #2}{%
  }{%
    \dimen@=\zref@extract{#3#2}{mzheight}\relax
    \ifdim\dimen@>\ht0 %
      \ht0=\dimen@
    \fi
    \dimen@=\zref@extract{#3#2}{mzdepth}\relax
    \ifdim\dimen@>\dp0 %
      \dp0=\dimen@
    \fi
  }%
  \copy0\relax
}
\makeatother

\usepackage{microtype}
\usepackage{graphicx}
\usepackage{booktabs} %

\usepackage{hyperref}

\usepackage[accepted]{icml2025}

\usepackage{amsmath}
\usepackage{amssymb}
\usepackage{mathtools}
\usepackage{amsthm}

\usepackage[capitalize,noabbrev]{cleveref}

\theoremstyle{plain}

\theoremstyle{definition}

\theoremstyle{remark}

\usepackage[textsize=tiny]{todonotes}

\usepackage{subcaption}  %

\usepackage{makecell}

\newcommand{\archshort}{DRAW}
\newcommand{\methodshort}{ReDRAW}

\newcommand{\shortparagraph}[1]{%
  \vspace{-1.5ex} %
  \paragraph{#1}%
}

\usepackage{dsfont}

\icmltitlerunning{Adapting World Models with Latent-State Dynamics Residuals}

\begin{document}

\twocolumn[
\icmltitle{Adapting World Models with Latent-State Dynamics Residuals}

\icmlsetsymbol{equal}{*}
\icmlsetsymbol{alumni}{\textdagger}

\begin{icmlauthorlist}
\icmlauthor{JB Lanier}{uci}
\icmlauthor{Kyungmin Kim}{uci}
\icmlauthor{Armin Karamzade}{uci}
\icmlauthor{Yifei Liu}{alumni}
\icmlauthor{Ankita Sinha}{alumni}
\icmlauthor{Kat He}{alumni}
\icmlauthor{Davide Corsi}{uci}
\icmlauthor{Roy Fox}{uci}
\end{icmlauthorlist}

\icmlaffiliation{uci}{Department of Computer Science, University of California Irvine. \textsuperscript{\textdagger}Work done while at UCI.}

\icmlcorrespondingauthor{JB Lanier}{jblanier@uci.edu}

\icmlkeywords{Machine Learning, ICML}

\vskip 0.3in
]

\printAffiliationsAndNotice{}  %

\begin{abstract}

Simulation-to-reality reinforcement learning (RL) faces the critical challenge of reconciling discrepancies between simulated and real-world dynamics, which can severely degrade agent performance. A promising approach involves learning corrections to simulator forward dynamics represented as a residual error function, however this operation is impractical with high-dimensional states such as images. To overcome this, we propose \methodshort{}, a latent-state autoregressive world model pretrained in simulation and calibrated to target environments through residual corrections of latent-state dynamics rather than of explicit observed states. Using this adapted world model, \methodshort{} enables RL agents to be optimized with imagined rollouts under corrected dynamics and then deployed in the real world. In multiple vision-based MuJoCo domains and a physical robot visual lane-following task, \methodshort{} effectively models changes to dynamics and avoids overfitting in low data regimes where traditional transfer methods fail.

\end{abstract}

\begin{figure*}[t]
   \centering
   \includegraphics[width=1.0\textwidth]{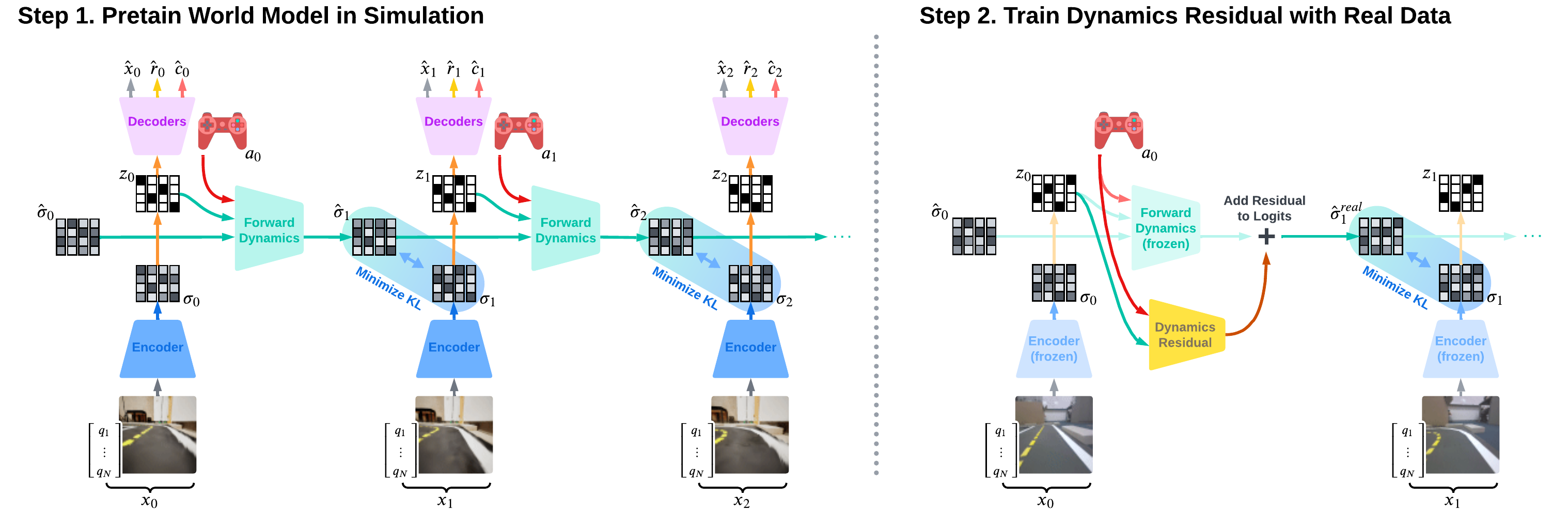}
   \vspace{-20pt}
   \caption{(\textbf{Left}) The \archshort{} world model is trained to encode states into a latent representation, from which states, rewards, terminations, and future latent states are predicted. An RL agent is trained in the world model via synthetic rollouts. 
    (\textbf{Right}) World model dynamics can be calibrated to a target environment by training a residual error correction on latent state dynamics predictions, allowing the RL agent to be trained under rectified dynamics.}
   \label{fig:simdreamer_hero_figure}
\end{figure*}

\section{Introduction}

Training robot control policies with reinforcement learning~(RL) in real-world environments is inherently expensive, time-consuming, and risky because it requires extensive interactions with physical systems. Simulation provides a promising alternative as it offers a controlled, cost-effective, and parallelizable setting for generating data and training capable policies. However, leveraging simulated environments effectively is challenging due to inaccuracies in their representation of agent observations and dynamics. These inaccuracies create a \textbf{sim-to-real gap}, where simulated environments fail to capture every relevant detail of real-world physics. This gap arises when real-world dynamics are only partially understood or are too expensive to model accurately. As a result, agents trained in simulation often struggle to transfer their policies directly to real-world settings without additional adaptation~\cite{kaufmann2023champion}.

One approach to addressing this gap is to use a small amount of real-world data to learn corrections to simulated transition models, known as \textit{residual dynamics corrections}. These corrections adjust the simulated dynamics to better match real-world behavior, allowing for more accurate training of control policies \cite{kaufmann2023champion, schperberg2023real, golemo2018sim}. However, this approach relies on the ability to efficiently learn corrections, which is difficult when the state information is represented in high-dimensional formats such as images. In these cases, significant feature engineering is often required to extract compact and meaningful state representations for learning residuals.

This work introduces a novel method for learning residual dynamics corrections directly in the \textbf{latent state space} of world models, eliminating the need for explicit feature engineering. Specifically, we build on latent-state world models such as Dreamer \cite{dreamerv3} and TD-MPC~\cite{tdmpc2}, which encode high-dimensional observations into compact latent states. These latent states can then be used to predict future dynamics, rewards, policy values, and optimal actions. World models enable RL agents to plan and gather experience using synthetic trajectories in latent space, significantly reducing the need for real-world interactions.

We propose a new world-model architecture, \textbf{\archshort{}} (\textbf{D}ynamics-\textbf{R}esidual \textbf{A}daptable \textbf{W}orld model), that encodes observations into a \textbf{generalizable latent space} and supports efficient residual learning. After pretraining \archshort{} on simulated data, its weights are frozen to provide a stable latent representation. A small offline dataset of real-world trajectories is then used to learn a residual function in the latent space. This function corrects the world model's dynamics, enabling it to more accurately represent real-world behavior. We refer to this residual-calibrated model as \textbf{Re}ctified \archshort{} (\textbf{\methodshort{}}). RL agents can be trained with \methodshort{} using imagined rollouts, producing policies that perform well in the real environment. Importantly, we do not require reward labels from the real environment to make this calibration, extending \methodshort{}'s applicability to real scenarios where rewards are only available in simulation.

\newpage
We evaluate \methodshort{} on four vision-based DeepMind Control Suite (DMC) environments and further demonstrate the real-world usability of \methodshort{} in sim-to-real applications by adapting from simulation to a physical real-time visual-navigation task on the Duckiebot low-cost robot \citep{paull2017duckietown}. Our experimental results suggest that \methodshort{} outperforms traditional transfer learning methods in small data regimes to adapt to mismatched dynamics and avoids overfitting without early stopping. In real robot experiments, \methodshort{} successfully performs simulation-to-reality dynamics adaptation with only 10K real steps ($\sim$17-minute demonstration), transferring from simulation with synthetic visual inputs to real-world images collected on the robot. 

\subsection*{Contributions}
\begin{enumerate}
    \item We propose a new world-model architecture for dynamics adaptation in fully observable visual-control domains. \archshort{} encodes all state information into a single generalizable latent space suitable for transfer in low-data regimes.
    \item We demonstrate that the \methodshort{} architectural extension can learn residual corrections in the latent space of \archshort{} to efficiently transfer between domains with mismatched dynamics, using only a small amount of offline target-domain data without reward labels.
    \item We show that our method adapts dynamics from simulation to reality while also zero-shot transferring latent-state encoders from synthetic to real robot images.
    \item Additionally, we open-source the code for our Unreal Engine~\citep{unrealengine} Duckiebot visual-control simulator to help facilitate further sim-to-real transfer research. Code and videos are available at \url{https://redraw.jblanier.net}.
\end{enumerate}

\section{Related Work}

This research lies at the intersection of sim-to-real dynamics transfer and RL with latent-state world models.

\subsection{Transferring Dynamics with Explicit Representations}

 Sim-to-real transfer of dynamics aims to adapt existing simulators or dynamics models used for planning and policy optimization to better match real-world environments. One way to transfer dynamics from simulation to reality is to calibrate predefined simulator physics parameters to match the target environment, either directly from real data \citep{si2022grasp} or as a correction to existing parameters \citep{allevato2020tunenet}. 
 However, doing so can be insufficient if no good approximation of the real environment exists in the space spanned by the allowed range of these parameters. In such cases, a more expressive modification of the simulator state transition function may be needed. 
 
 Along these lines, \citet{ball2021augmented} and \citet{arcari2023bayesian} calibrate linear error models on simulator transition dynamics using real data for policy adaptation and, respectively, model predictive control. Similarly, \citet{mallasto2021affine} use affine transport to adapt simulator state dynamics models to real domains. 
 \citet{golemo2018sim} train an LSTM conditioned on state–action history to predict a state transition residual, and \citet{schperberg2023real} efficiently adapt a neural-network state-dynamics residual by using Unscented Kalman Fitlering. 
\citet{kaufmann2023champion} employ k-nearest neighbor regression and Gaussian process residuals on transition dynamics and state encodings to calibrate their simulator for drone racing at an expert human level.

Each of these methods relies on the assumption that the environment state can be represented with a compact vector representation with which a generalizable dynamics correction can be learned with a relatively low-complexity model and small real-data requirements.
We consider the case where the components of the state are instead in a \textit{high-dimensional} format like images and we do not have a predefined mapping from these states to such a necessary compact vector representation. To adapt simulation transition dynamics under these more difficult conditions, we propose to learn a latent-state world model of the simulation and then train a residual correction on the world model's dynamics to match transitions in the real environment.

\subsection{World Models with Latent State Spaces}

World models \citep{ha2018world} with latent state spaces are environment models in which planning and policy learning can be more efficient than with environment states due to a succinct representation of environment states and dynamics.
Dreamer \cite{dreamerv1, dreamerv2, daydreamer, dreamerv3} models environments in the stochastic POMDP~\cite{cassandra1998exact} by encoding observations as latent states and reconstructing future latent states, rewards, and observations. The Dreamer architecture allows agents to then train on synthetic experience by rolling out ``imagined" trajectories inside of the world model. TD-MPC \cite{tdmpc, tdmpc2} models deterministic fully observable MDPs by similarly reconstructing future latent states and rewards, as well as task value functions. 
TD-MPC2 \citep{tdmpc2} has shown good results learning shared features from a suite of environments to quickly transfer to new ones, while we focus on transferring from a single environment to a similar target environment by avoiding overfitting to limited data. 

Concerning exploration with world models, collecting diverse source trajectories was crucial in our experiments for learning transferable features and dynamics. To achieve this, we use Plan2Explore~\citep{sekar2020planning}, a method compatible with both Dreamer and our proposed \archshort{} architecture, which trains an auxiliary RL agent alongside the exploit policy to maximize model uncertainty in latent dynamics predictions, promoting wide-reaching exploration.

\subsection{Domain Randomization}

Domain randomization is widely used for sim-to-real transfer by exposing policies to diverse variations in images~\citep{tobin2017domain,james2019sim} or dynamics~\citep{peng2018sim,mehta2020active}.
However, training on a broad distribution of environment conditions can yield an overly conservative policy or require system identification at test time, potentially degrading test-time performance.

We apply limited camera-parameter randomization in Duckiebots to address perception gaps but focus on adapting to unforeseen dynamics. Rather than relying solely on randomization, we use a residual method to correct dynamics mismatches, achieving more flexible adaptation beyond pre-defined perturbations.

\section{Preliminaries and Problem Definition}
In this work, we consider two Markov Decision Processes~(MDPs), denoted as $M_{\text{sim}}$ and $M_{\text{real}}$, which share the same state space, action space, and reward function, but differ in their transition dynamics. Formally, each MDP is defined by a tuple $M_i = (X, A, R, \gamma, P_i)$, with a shared state space $X$, action space $A$, reward function upon entering a state $R: X \rightarrow \mathbb{R}$,  discount factor $\gamma \in [0, 1)$, and stochastic transition function $P_i$ for $i \in \{\text{sim}, \text{real}\}$. %

Our objective is to find a policy $\pi_\text{real}$ that achieves good expected discounted cumulative reward in $M_{\text{real}}$, $J_{\pi,\text{real}} = \mathbb{E}_{\pi, P_{\text{real}}} \left[ \sum_{t=0}^{\infty} \gamma^{t} R(x_t) \right]$. To capture logistic challenges common in real robot settings, we have access to a limited amount of offline reward-free data $(x_t, a_t, x_{t+1})$ from $M_{\text{real}}$, and to make up for this, we can collect a large amount of online experience $(x_t, a_t, x_{t+1}, r_{t+1})$ in $M_{\text{sim}}$.

Our work addresses the problem of rectifying one-step dynamics predictions to align them with actual outcomes in the real environment. In this work, we consider MDPs where all of the information necessary to predict future dynamics and rewards can be extracted from the agent's immediate observation of the world state, $x_t$. A world model will then encode $x_t$ into a latent state $z_t$, which in an MDP can also be a state of the latent dynamics. To maintain this assumption of full observability in our experiments on image-based domains, we supplement visual inputs with vectors of otherwise absent complementary information such as velocity values. We leave relaxing \methodshort{}'s limitation to MDPs only and addressing the partially observable case for future research.

Additionally, our proposed residual method assumes that only transition dynamics vary between $M_{\text{sim}}$ and $M_{\text{real}}$, and that the agent's perceptual processing itself can effectively transfer zero-shot between the two environments. Independent of our latent dynamics residual method, we demonstrate that it is possible to overcome image distribution shifts with world models between our Duckiebot simulation and real robot domains by using image augmentations with an asymmetric state encoder/decoder objective similar to \citet{kim2024make} and a digital-twin simulator that leverages Gaussian splatting \citep{gaussiansplatting} with mild visual domain randomization.
We note that the assumption of a shared state space does not lose generality, in principle, because the dynamics can be modeled as leading to state subspaces that are disjoint between the environments; but in practice, visual transfer does help to keep the residual simple.

\section{Method}

In this section, we describe our MDP world model architecture \archshort{} (Figure~\ref{fig:simdreamer_hero_figure}, Left) and its counterpart with calibrated dynamics, \methodshort{} (Figure~\ref{fig:simdreamer_hero_figure}, Right). We first define the \archshort{} model, how it represents latent states and dynamics, and how it is trained. Then we describe how we facilitate sample efficient transfer learning of dynamics by training a residual error correction on latent-state transitions, creating the \methodshort{} world model.

\subsection{\archshort{} Architecture and Pretraining}

We use \archshort{} to model an MDP by encoding state inputs into a compressed stochastic latent representation using variational inference. Similar to DreamerV3 \citep{dreamerv3}, our latent representation is trained via objectives for state and reward reconstruction along with future latent-state prediction. We then train an actor–critic reinforcement-learning agent on latent-state inputs by autoregressively rolling out synthetic trajectories as experience and using reconstructed rewards as a learning signal. Finally, the actor can be deployed to the environment by encoding immediate state inputs as latent states and providing these encodings to the actor. Figure~\ref{fig:simdreamer_hero_figure} (Left) depicts connections during \archshort{} world model training, while Figure~\ref{fig:actor_critic_training_and_deployment} in Appendix~\ref{sec:actor_critic_training} depicts actor–critic training and deployment.

We model the latent state purely as a single stochastic multi-categorical discrete variable $z_t \in \mathcal{Z}$. $z_t$ is a $K$-tuple of conditionally independent categorical variables, each represented as a 1-hot vector of length $N$. We denote $z_t$ as the latent state encoded from the immediate state $x_t$~\eqref{eq:state_encoder} and $\hat{z}_t$ as the latent state predicted via world model dynamics from the previous latent state and action~\eqref{eq:forward_sample}. We denote $\hat{u}_t \in \mathbb{R}^{K\times N}$ as the logits for the multi-categorical distribution of $\hat{z}_t$ and $\hat{\sigma}_t = \R{softmax}(\hat{u}_t)$ as the $K$ concatenated normalized probability vectors. To estimate gradients in the sampling step for $z_t$ or $\hat{z}_t$, we use the straight-through estimator \citep{bengio2013estimating, dreamerv2}.

By compressing all state information into a single discrete representation \(z_t\), 
we aim to provide a well-structured encoding of the underlying state \(x_t\), 
enabling the learning of generalizable functions, such as residual corrections,
from limited data using \(z_t\) as input.
As illustrated in Figure~\ref{fig:simdreamer_hero_figure}~(Left), we define our \archshort{} world model and actor–critic RL agent, respectively parameterized by $\theta$ and $\phi$, as follows:

\allowdisplaybreaks

\begin{alignat}{3}
    \text{State Encoder}\quad& &z_t& & &\sim q_\theta(z_t | x_t) \label{eq:state_encoder} \\
    \text{Forward Dynamics}\quad& &\hat{u}_t& & &= f_\theta(z_{t-1}, \hat{\sigma}_{t-1}, a_{t-1}) \\
    \text{Forward Belief}\quad& &\hat{\sigma}_t& & &= p_\theta(\hat{z}_t | z_{t-1}, \hat{\sigma}_{t-1}, a_{t-1}) \label{eq:forward_belief} \\
    & & & & &= \R{softmax}(\hat{u}_t) \\
    \text{Forward Sample}\quad& &\hat{z}_t& & &\sim \mathrm{MultiCategorical}(\hat{\sigma}_t) \label{eq:forward_sample}\\
    \text{Reward}\quad& &\hat{r}_t& & &\sim p_\theta(\hat{r}_t | z_t) \\
    \text{Continue}\quad& &\hat{c}_t& & &\sim p_\theta(\hat{c}_t | z_t) \\
    \text{State Decoder}\quad& &\hat{x}_t& & &\sim p_\theta(\hat{x}_t | z_t) \\
    \text{Policy}\quad& &a_t& & &\sim \pi_\phi(a_t | z_t) \\
    \text{Value Function}\quad& &v_t& & &= V_\phi^\pi(z_t).
\end{alignat}

We represent all functions in \archshort{} as multi-layer perceptrons (MLPs) except for image components of the state encoder and decoder, which are convolutional (CNNs). Interestingly, we found that providing $\hat{\sigma}_t$ as an input to the forward dynamics function $f_\theta$ significantly increased our downstream adaptation performance. We speculate that this is because $\hat{\sigma}_t$ helps provide a gradient signal for learning features relevant for long-term dynamics predictions without adding additional dimensionality to state prediction outputs. We provide ablations on this design choice in Appendix~\ref{sec:architecture_ablations}.

We optimize \archshort{} on $M_\textit{sim}$ with a prediction loss $\mathcal{L}_\text{pred}$ to reconstruct states, rewards, and terminations, as well as a dynamics loss $\mathcal{L}_\text{dyn}$ and a representation loss $\mathcal{L}_\text{rep}$ to learn latent-state dynamics under a predicable representation. Drawing subtrajectories $\zeta$ from a buffer of interaction experience, the world-model loss function $\C{L}(\theta)$ is:
\eqn{
\MoveEqLeft \C{L}(\theta) = \E_{q_\theta (z_{1:T}|\zeta)} \Bigg[\sum_{t=1}^{T} \beta_\text{pred}\mathcal{L}_\text{pred}^t(\theta) \nn\\&+
    \beta_\text{dyn}\mathcal{L}_\text{dyn}^t(\theta) +
    \beta_\text{rep}\mathcal{L}_\text{rep}^t(\theta)\Bigg],
}

where $T$ is the length of $\zeta$, and for $t = 1, \ldots, T$:
\eqn{
    \mathcal{L}_\text{pred}^t(\theta) \dot{=}& -\ln{p_\theta(x_t|z_t)} - \ln{p_\theta(r_t|z_t)} - \ln{p_\theta(c_t|z_t)} \\
    \mathcal{L}_\text{dyn}^t(\theta) \dot{=}& [\D[q_{\bar{\theta}}(z_t|x_t) || p_\theta(\hat{z}_t|z_{t-1}, \hat{\sigma}_{t-1},a_{t-1})]]_1 \label{dyn_loss}\\
    \mathcal{L}_\text{rep}^t(\theta) \dot{=}& [\D[q_\theta(z_t|x_t) || p_{\bar{\theta}}(\hat{z}_t|z_{t-1}, \hat{\sigma}_{t-1},a_{t-1}))]]_1,
}
with $[\cdot]_1$ denoting clipping to 1 any value below 1, $\D$ the Kullback–Leibler divergence, and $\bar{\theta}$ a stopped-gradient copy of $\theta$.

We train the actor–critic agent with the same procedure and losses as DreamerV3, providing the \archshort{} world model state $\hat{z}_t$ as agent inputs during imagined rollouts and $z_t$ during evaluation. When training the actor–critic, we seed synthetic rollouts with starting states $x_0$ drawn from the same experience buffer as used for world-model training. We do not backpropagate value gradients through dynamics, and we train the policy using the Reinforce objective~\citep{williams1992simple} with normalized returns and critic baselines~\citep{dreamerv3}.

We alternate mini-batch updates between the world model and the actor–critic.
During source environment pretraining, updates are interleaved with online data collection. 
Since we cannot fully predict which trajectories in $M_\textit{sim}$ will best facilitate 
learning transferable features and dynamics for $M_\textit{real}$, we employ Plan2Explore~\cite{sekar2020planning} 
to provide intrinsically motivated exploration, ensuring that we collect a large and highly diverse set of source-environment trajectories.

\begin{figure*}[h]
   \centering
   \includegraphics[width=1.0\textwidth]{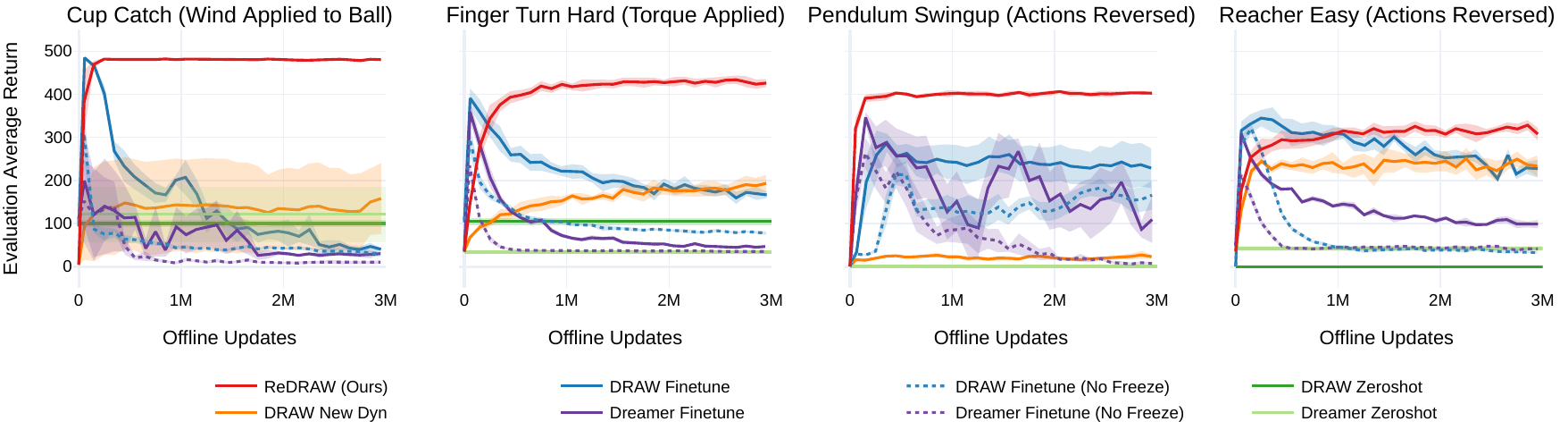}
\vspace{-20pt}
   \caption{Average evaluation episode return transferring from each DMC environment to a modified variant of it given $4e4$ offline target environment transition samples. Shaded regions indicate the standard error of the mean over 4 seeds for each method. \methodshort{} consistently achieves high returns in the target environments and avoids overfitting.}
   \label{fig:main_result}
\end{figure*}

\subsection{Adaptation via Latent Dynamics Residuals}
After pretraining the \archshort{} world model online in the $M_\textit{sim}$ environment with a large amount of data, we propose the \methodshort{} architecture and method to use a small offline dataset of transitions from the target environment to calibrate \archshort{}'s dynamics to match $M_\textit{real}$ using a latent-state error residual.

We model the dynamics residual using an MLP $\delta_\psi$ that predicts a correction $\hat{e}_t $ to the forward-dynamics logit vector $\hat{u}_t$. This correction produces a modified transition distribution $\hat{\sigma}_t^{\textit{real}}$, from which forward latent-state predictions $\hat{z}_t^\textit{real}$ are sampled to approximate $M_\textit{real}$. That is, we model the calibrated dynamics as:
\begin{align}
       \hat{u}_t &= f_\theta(z_{t-1}, \hat{\sigma}^\textit{real}_{t-1}, a_{t-1}) \\
       \hat{e}_t &= \delta_\psi(z_{t-1}, a_{t-1}) \\
       \hat{\sigma}_t^{\textit{real}} &= p_{\theta,\psi}(\hat{z}^\textit{real}_t|z_{t-1}, \hat{\sigma}_{t-1}^\textit{real}, a_{t-1}) \\
    &= \R{softmax}(\hat{u}_t + \hat{e}_t) \\
       \hat{z}_t^\textit{real} &\sim \R{MultiCategorical}(\hat{\sigma}_t^{\textit{real}}).
\end{align}

To train the residual on real data, we freeze the world-model weights $\theta$ and only optimize the parameters \( \psi \) of the residual network \( \delta_\psi \). In the transfer phase, we only optimize the actor–critic agent and a new loss $\mathcal{L}_\delta(\psi)$ on the rectified world-model dynamics. Our objective is to predict corrections $\hat{e}_t$ of $\hat{u}_t$ so that our new dynamics predictions \( \hat{\sigma}_t^{\textit{real}} \) match the observed encoder distribution over latent states collected in $M_{\text{real}}$. The loss function for the residual is:

\eqn{
    \label{eq:residual_loss}
   \MoveEqLeft \mathcal{L}_\delta(\psi) = \mathbb{E}_{q_{\bar{\theta}} (z_{1:T}|\zeta^\textit{real})} 
   \Bigg[\sum_{t=1}^{T} \D[q_{\bar{\theta}}(z_t|x_t) || \nn\\
   & p_{\bar{\theta},\psi}(\hat{z}^\textit{real}_t|z_{t-1}, \sigma_{t-1}^\textit{real}, a_{t-1})]\Bigg].
}

Because we consider fully observable environments, the target encoder latent-state distribution $q_\theta(z_t|x_t)$ depends solely on $x_t$ and can be frozen after pretraining in $M_{\text{sim}}$ if the collected source-environment data adequately covers the state space.

Finally, given the \methodshort{} world model with dynamics adapted to match $M_{\text{real}}$, the actor–critic can learn a high-performing policy for the new environment by training in the world model under the new rectified dynamics, using $\hat{z}_t^\textit{real}$ as input during training. In our experiments, we still alternate agent and world-model training during adaptation, although in principle, the agent could be trained after world-model training is stopped. 

Notably, the latent-state representation for \methodshort{} in $M_{\text{real}}$ is unchanged from \archshort{} in $M_{\text{sim}}$. As a result, the frozen \archshort{} $M_{\text{sim}}$ reward function $p_\theta(\hat{r}_t | \hat{z}_t)$ can be reused in world-model rollouts to train the \methodshort{} agent with $p_\theta(\hat{r}_t | \hat{z}^\textit{real}_t)$ in $M_{\text{real}}$,  eliminating the need for reward data from $M_{\text{real}}$. This is particularly beneficial since setting up a reward recording system in real-world scenarios, such as robotics, often requires costly and complex setups like additional sensors or feedback mechanisms, which may be infeasible in certain environments.

\section{Experiments}
We evaluate \methodshort{} in two distinct settings: (1) adapting from DeepMind Control (DMC)~\cite{tassa2018deepmind} environments to modified counterparts with changed physics, and (2) transferring from a simulation in Unreal Engine to a real robot visual lane-following task using the Duckietown platform. Our experiments address three main questions:

\begin{enumerate}
    \item How do latent-space residuals compare to traditional finetuning methods in correcting world-model dynamics under limited target-domain data?
    \item How do data quantity and collection policies influence transfer performance?
    \item Can \methodshort{} effectively close the sim-to-real gap in a physical robotics task with visual inputs?
\end{enumerate}

\subsection{DeepMind Control Experiments}

\subsubsection{DMC Domains}
We first consider four pairs of source and target environments from the DMC suite, each pair having the same state and reward structure but mismatched dynamics. We use original environments from DMC as sources, while the target environments introduce physics modifications such as applied wind, external torque, or reversed actions. For a detailed description, refer to Appendix \ref{sec:appendix:dmc_exp}.
These differences in dynamics between source and target environments are substantial enough to require policy adaptation for optimal performance. Although dynamics differ between source and target, the state spaces, reward functions, and termination conditions remain unchanged.

To pretrain on each source environment, we collect 9 million environment steps (4.5e6 decisions steps with an action repeat of 2) using Plan2Explore \citep{sekar2020planning}, which promotes diverse state visitation rather than narrowly exploiting the original environment’s reward function. After this phase, we adapt to each target environment using a small offline dataset of 40K decision steps (equivalent to 80 episodes), gathered by an expert policy in the target domain.

\subsubsection{Comparison with Finetuning}

We compare \methodshort{} with several baselines that attempt to adapt a pretrained world model to the new domain. Critically, except where noted with \textbf{*}, the methods we test do not use reward labels or train with a reward-reconstruction objective during the adaptation phase. These baselines include:

\shortparagraph{\archshort{}/DreamerV3 Zeroshot:} We take the source-trained \archshort{} or DreamerV3  agent and deploy it in the target environment without any adaptation.

\shortparagraph{\archshort/DreamerV3 Finetune:} The world model and agent are finetuned on the target domain offline dataset. To mitigate overfitting on the small dataset, we freeze the world model encoder and decoder parameters and only retrain the agent and dynamics components. For \archshort{} this entails optimizing only $f_\theta$ with $\mathcal{L}_\text{dyn}$. Analogously, for DreamerV3, the RSSM recurrent prior and posterior components are updated while leaving the observation feature embeddings and decoders unchanged.

\shortparagraph{\archshort{}/DreamerV3 Finetune (No Freeze)*:} Every component, including the encoder and decoder, is finetuned using all original world model loss terms. These are the only two baselines requiring access to reward data during adaptation.

\shortparagraph{\archshort{}{} New Dyn: } The entire world model is frozen after pretraining, but instead of learning a residual addition to $f_\theta$, we train a new dynamics function $\hat{\sigma}_t^\textit{real} = g_\psi(\hat{\sigma}_{t}, z_{t-1}, a_{t-1})$ conditioned on the distribution predicted by the frozen source dynamics (Eq.~\ref{eq:forward_belief}). This ablation demonstrates an alternate way to leverage frozen dynamics predictions learned from the source environment. Other variations of this baseline are compared in Appendix~\ref{sec:new_dyn_fn}.

\begin{figure*}[h]
   \centering
   \includegraphics[width=1.0\textwidth]{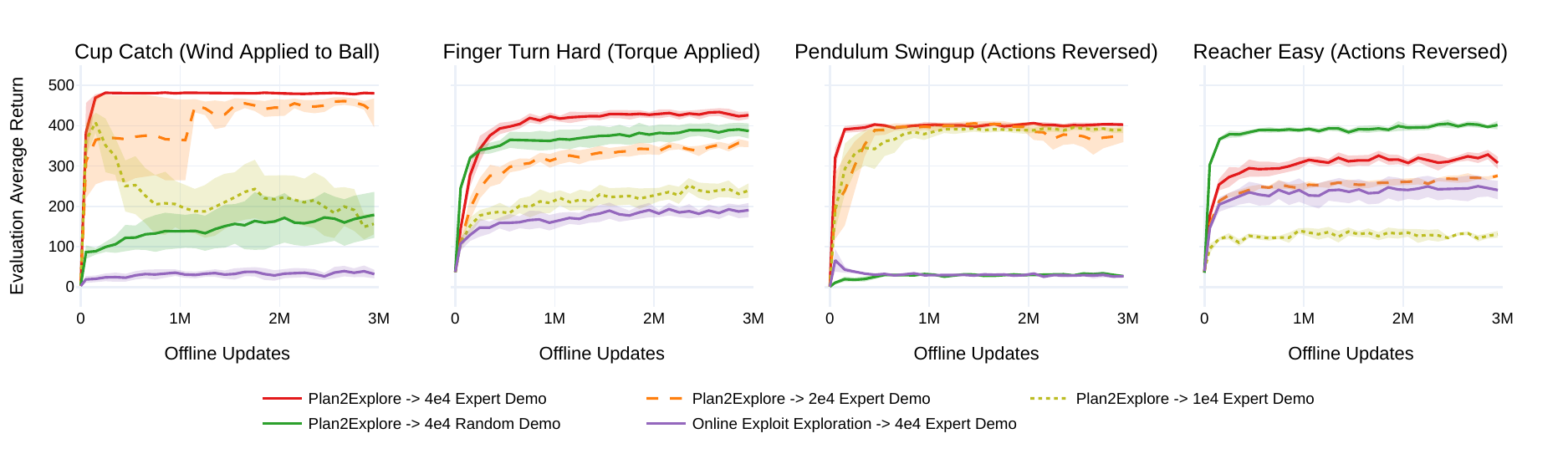}
\vspace{-20pt}
   \caption{Impact of offline adaptation dataset size and source/target domain data collection strategies on \methodshort. Expert demonstrations consistently provide useful target domain data for adaptation. Collecting diverse simulation experience with a method like Plan2Explore is essential for good transfer performance.}
   \label{fig:data_combined_ablation}
\end{figure*}

Figure~\ref{fig:main_result} shows the returns in each target domain as a function of offline updates on each target dataset. Direct deployment without adaptation fares poorly in these altered dynamics. Finetuning approaches initially improve in some cases but all eventually overfit to the small dataset. The \emph{No Freeze*} variations are quicker to overfit than their partially frozen counterparts. In contrast, \methodshort{}'s latent-space residual method attains a sustained level of high-performance and avoids overfitting during the 3 million updates (1-3 days of training) we test on. 
This highlights a critical benefit of the \methodshort{} transfer method: once \methodshort{} reaches high performance in the target domain, it demonstrates a remarkable resistance to performance degradation. \methodshort{}'s ability to avoid overfitting for long periods of time makes it highly applicable to sim to real scenarios where validation testing on a real robot often cannot practically be done repeatedly and educated guesses need to be reliably made regarding stopping conditions. 

\methodshort{} excels at maintaining a high degree of validation performance by preserving existing dynamics predictions learned in simulation where data is abundant and using the limited target data to learn a low-complexity adjustment to those predictions. Comparing \methodshort{} with \emph{\archshort{} New Dyn}, we see that while both approaches utilize both the previous state and the frozen simulation dynamics predictions, the residual operation appears to play a key role in limiting the complexity of the changes made to the original world model dynamics, allowing \methodshort{} to avoid overfitting.

\subsubsection{Data Policies and Quantity}

Figure~\ref{fig:data_combined_ablation} examines how pretraining and target-domain data collection methods influence \methodshort{}'s performance. In our default scenario, we pretrain in the source domain with Plan2Explore, and then adapt to 40K steps of expert demonstrations in the target environment. Alternatively, we test gathering source data with the training exploit policy (rather than Plan2Explore) and replacing the expert dataset with  random actions in the target domain. While expert trajectories reliably lead to effective adaptation, random actions alone only occasionally suffice, suggesting that a mixture of expert data and high entropy actions may help the agent learn counterfactuals and avoid suboptimal outcomes.

We further see that reducing the number of expert demonstrations from 40K to 10K can cause \methodshort{} to overfit, particularly in the Cup Catch environment. On the pretraining side, collecting a broad range of trajectories with Plan2Explore proves beneficial. When we collect data with a policy optimized only for source environment, the necessary transitions for the target domain’s altered dynamics may be missed. Because it is difficult to predict which states and actions will matter under new physics, gathering diverse pretraining data is likely essential for reliable transfer.

\begin{figure*}[]
  \centering
  \newcommand{\imgheight}{2.8cm} %

  \begin{subfigure}{0.29\textwidth}
    \centering
    \includegraphics[height=\imgheight]{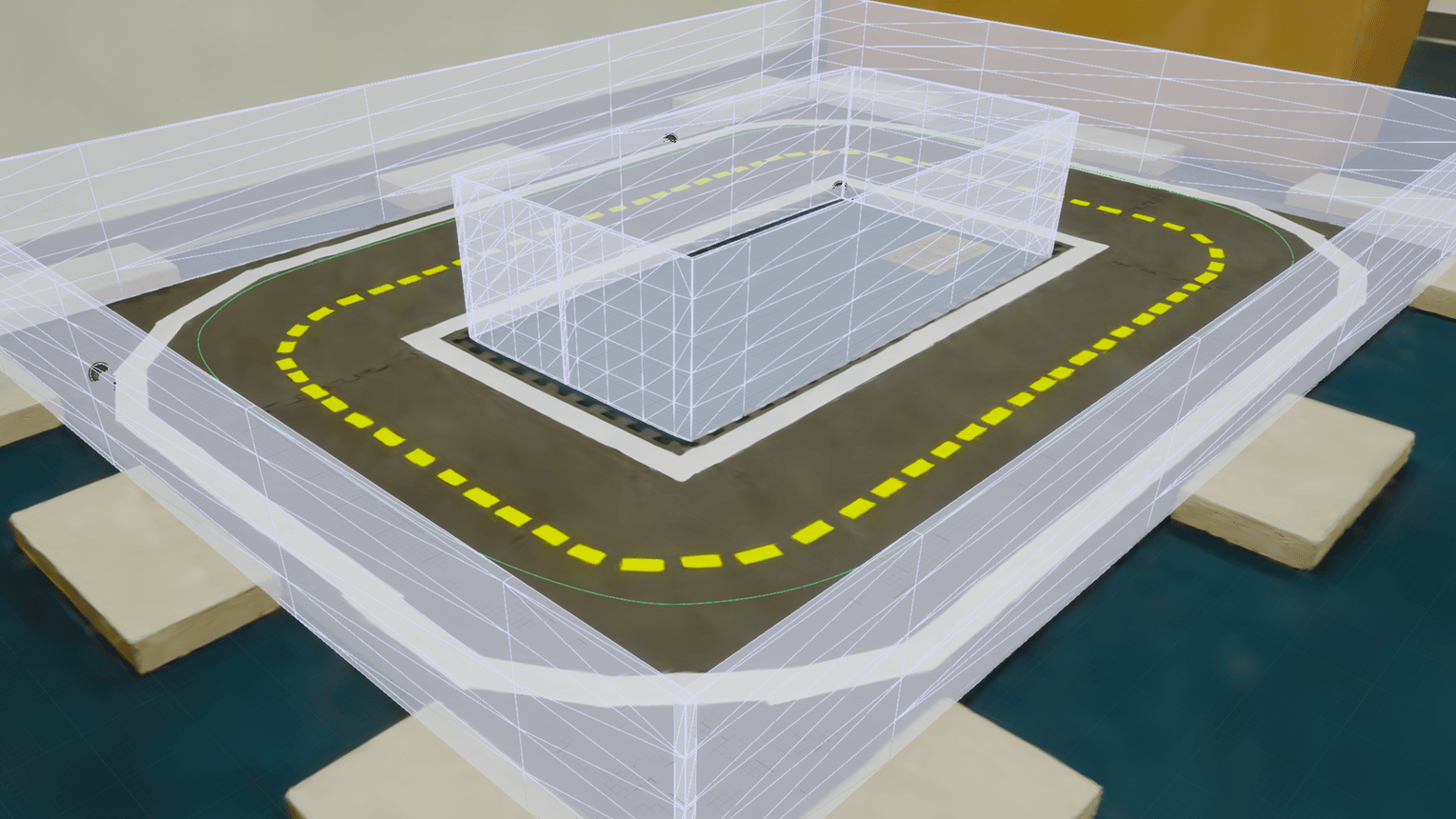}
    \caption{Digital Twin Simulation}
    \label{fig:duckie_twin_overview}
  \end{subfigure}
  \hspace{3pt} %
  \begin{subfigure}{0.29\textwidth}
    \centering
    \includegraphics[height=\imgheight]{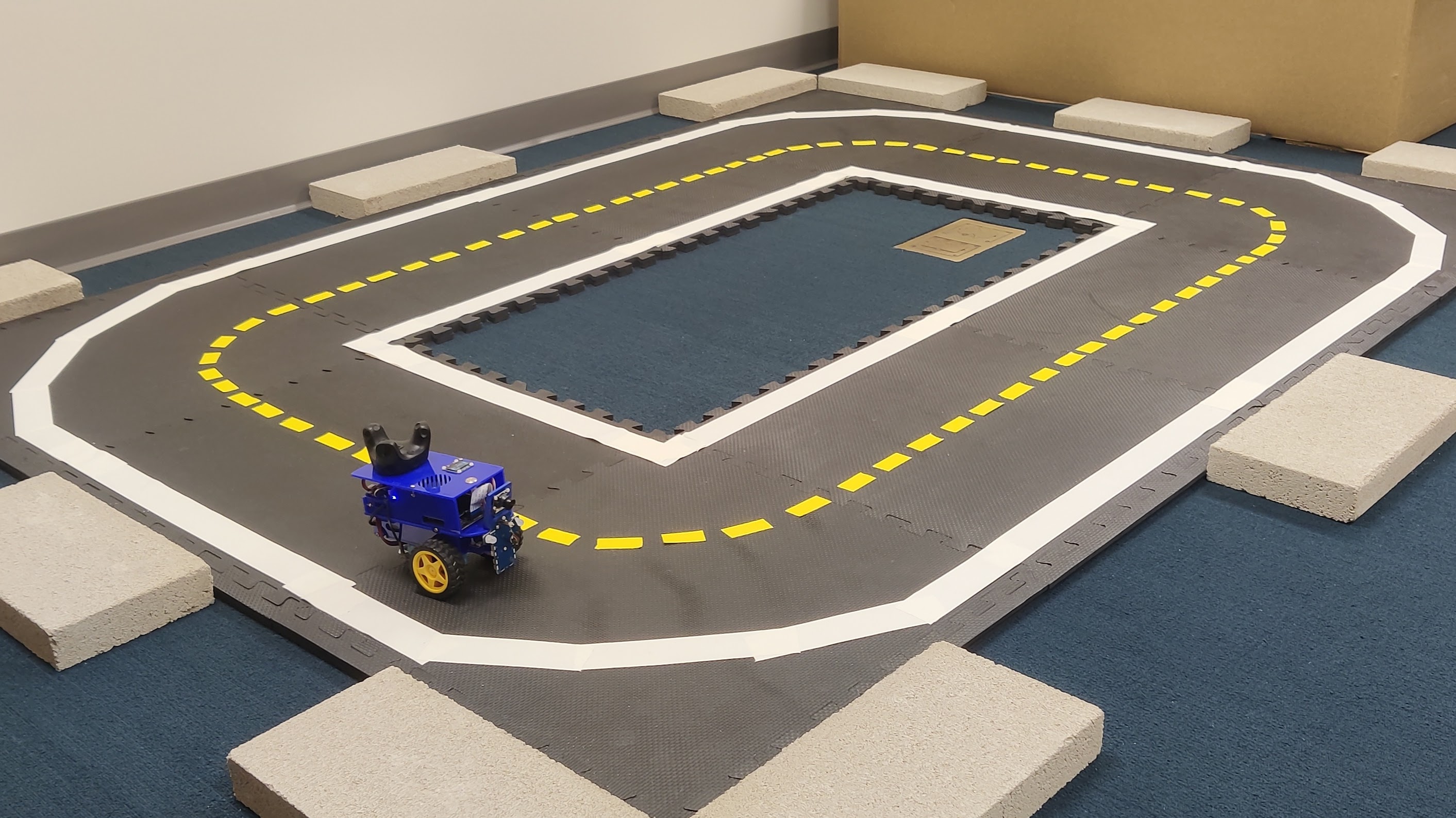}
    \caption{Real Robot Environment}
    \label{fig:duckie_real_overview}
  \end{subfigure}
  \hspace{3pt} %
  \begin{subfigure}{0.16\textwidth}
    \centering
    \includegraphics[height=\imgheight]{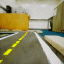}
    \caption{Sim State Image}
    \label{fig:duckie_twin_obs}
  \end{subfigure}
  \hspace{3pt} %
  \begin{subfigure}{0.16\textwidth}
    \centering
    \includegraphics[height=\imgheight]{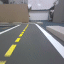}
    \caption{Real State Image}
    \label{fig:duckie_real_obs}
  \end{subfigure}
\vspace{-10pt}
  \caption{(a) Digital-twin simulation constructed using Gaussian splatting \cite{gaussiansplatting}. (b) Real-world robot lane-following environment. (c) Simulation state image component. (d) Real-world state image component. The agent is tasked to drive quickly around the track while staying near the lane center using an egocentric camera and velocity sensor. We train our \archshort{} world model in simulation and calibrate its dynamics with \methodshort{} on offline real trajectories, producing a successful agent in the real environment.
  }
  \label{fig:duckiebot}
\end{figure*}

\begin{table}[h]
   \centering
   \setlength{\tabcolsep}{1.6pt} 
   \caption{Mean and SEM performance on the real Duckiebots lane-following task aggregated over 5 evaluation episodes each for 4 training seeds. Agents are given 300 steps to complete a lap from a fixed starting position. \emph{Center Offset} denotes distance from the lane center. Absence of a \emph{Lap Time} indicates all runs either failing to complete a lap or terminating early by driving off the track.}
   \label{table:duckiebots_results}

   \begin{tabular}{lrrr}
      \toprule
      \textbf{Method} 
        & \textbf{\thead{Avg Dense \\Reward ($\uparrow$)}} 
        & \textbf{\thead{Avg Lap \\ Time ($\downarrow$)}} 
        & \textbf{\thead{Avg Center \\ Offset ($\downarrow$)}} \\
      \midrule
      \multicolumn{4}{l}{\textbf{Transfer Sim to Unmodified Real}} \\
      \midrule
      \methodshort{} (Ours) 
        & \textbf{0.38} {\footnotesize $\pm$ 0.02} 
        & \textbf{22.75} {\footnotesize $\pm$ 0.75}
        & \textbf{2.39} {\footnotesize $\pm$ 0.15} \\

      \archshort{} Zeroshot 
        & 0.12 {\footnotesize $\pm$ 0.03} 
        & \textbf{22.41} {\footnotesize $\pm$ 0.73} 
        & 4.79 {\footnotesize $\pm$ 0.20} \\

      Dreamer Finetune 
        & -0.87 {\footnotesize $\pm$ 0.31}
        & \ccell - 
        & 5.45 {\footnotesize $\pm$ 1.10} \\

      Dreamer Zeroshot 
        & -1.18 {\footnotesize $\pm$ 0.13}
        & \ccell - 
        & 6.86 {\footnotesize $\pm$ 0.45} \\
      \midrule
      \multicolumn{4}{l}{\textbf{Transfer Sim to Actions Reversed Real}} \\
      \midrule
      \methodshort{} (Ours) 
        & \textbf{0.40} {\footnotesize $\pm$ 0.01} 
        & \textbf{24.21} {\footnotesize $\pm$ 0.72} 
        & \textbf{2.07} {\footnotesize $\pm$ 0.15} \\

      \archshort{} Zeroshot 
        & -2.72 {\footnotesize $\pm$ 0.17} 
        & \ccell - 
        & 9.39 {\footnotesize $\pm$ 0.68} \\

      Dreamer Finetune 
        & -1.61 {\footnotesize $\pm$ 0.27}
        & \ccell - 
        & 7.75 {\footnotesize $\pm$ 0.98} \\

      Dreamer Zeroshot 
        & -2.35 {\footnotesize $\pm$ 0.27} 
        & \ccell - 
        & 13.36 {\footnotesize $\pm$ 0.82} \\
      \bottomrule
   \end{tabular}
\end{table}

\subsection{Duckiebot Sim-to-Real Transfer}

\subsubsection{Task and Setup}
Finally, we evaluate \methodshort{} in a sim-to-real robotic lane-following task using the Duckietown platform \cite{paull2017duckietown}. Here, the agent controls a wheeled robot to navigate around a track while remaining centered in its lane. The state space includes a forward-facing camera image plus egocentric forward and yaw velocity values, and actions defined as continuous forward and yaw target velocities in [-1, 1].

To provide a simulation to transfer from, we construct an environment in Unreal Engine using a Gaussian splat \cite{gaussiansplatting} reconstruction of the robot's environment to mimic the robot's state space. Figure~\ref{fig:duckie_twin_overview} and \ref{fig:duckie_real_overview} show the digital twin and real environment, respectively. We also implement the simulation with a rough approximation of real dynamics, although details like precise handling while driving and control rate (6Hz sim vs 10Hz real) still differ from the real robot.  

The Duckiebot receives rewards proportional to its projected velocity along the lane-center path but instead incurs penalties when it deviates too far from this path. When moving forward, we also penalize the agent proportionally its yaw velocity to encourage smooth driving. Episodes terminate either when the robot leaves the track, with a large penalty applied, or after 200 steps. Exact experiment details are presented in Appendix~\ref{sec:duckiebot_details}.

We provide the agent with reward data during simulation pretraining, and we do not provide reward labels in training data collected from the real environment. In order to measure test-time deployment performance, during real evaluation only, we record the robot's location with an HTC Vive motion tracker to measure equivalent simulation rewards, lap times, and the robot's distance from the lane center.

\subsubsection{Bridging the Sim-to-Real Vision Gap}
Despite efforts to recreate the real environment, visual disparities between the simulation and real environment still exist (Figure~\ref{fig:duckie_twin_obs} vs. \ref{fig:duckie_real_obs}). Although our main focus in this paper is adapting dynamics, we employ mild per-episode domain randomization~\cite{tobin2017domain} along with image augmentation~\cite{info11020125} to bridge the sim-to-real vision gap. We domain-randomize the simulation camera's mounted location on the robot, camera tilt, and its field of view. We also apply image augmentations at train time to both sim and real image inputs to learn world model image encoders robust to task-irrelevant features like lighting, color hue, and lab furniture placement. Although we train with augmented inputs, our decoder reconstructs the original images as targets in $\mathcal{L}_\text{pred}$, thus focusing the latent-space features on task-relevant elements rather than the irrelevant augmentations we apply. We apply this asymmetric decoding objective to both \archshort{} and DreamerV3. As described above, \methodshort{} trains its residual with augmented inputs but has no decoding objective during transfer learning.

\subsubsection{Transferring to the Real Robot}
We pretrain \archshort{} and DreamerV3 in simulation using 600K random actions followed by 1.4 million online steps with Plan2Explore. 
On the real robot, we collect a small offline adaptation dataset of 1e4 timesteps ($\sim$17 minutes) using human demonstrations employing a mixture of proficient driving and random safe actions. Table~\ref{table:duckiebots_results} compares performance using this offline dataset to adapt to two variations of the real environment, \emph{unmodified real} where minor physics disparities between sim and real are the natural result of inaccurate dynamics modeling, and \emph{actions-reversed real}, where actions (in adaptation data and deployment) are inverted, requiring large but regular adaptation to drive successfully. We adapt \methodshort{ and} DreamerV3 \emph{Finetune} for 2e5 offline updates. We evaluate 5 episodes attempting a lap on the robot each for 4 training seeds. 

In \emph{unmodified real}, \archshort{} \emph{zeroshot} is able to successfully drive despite never seeing real data but incurs low rewards by veering far from the lane center. DreamerV3 \emph{zeroshot} fails, driving off the track in all lap attempts. We speculate that DreamerV3 \emph{zeroshot} fails while \archshort{} \emph{zeroshot} succeeds because DreamerV3's recurrent model observes OOD sequences under changed dynamics, resulting in inaccurate latent-state predictions. \archshort{} and \methodshort{} are non-recurrent in deployment time and cannot suffer from this same issue. Similar to DMC experiments, DreamerV3 \emph{Finetune} fails to adapt, possibly due to overfitting, and \methodshort{} achieves significantly higher average dense rewards than \archshort{} \emph{zeroshot} by training with corrected dynamics and staying close to the lane center.

In the more extreme \emph{actions-reversed real} transfer task, \methodshort{} is the only method that successfully adapts and completes laps on the real robot due to the incompatibility of zero-shot policies to this environment and the limited real data in the case of DreamerV3 \emph{Finetune}. These results demonstrate that \methodshort{} can be effectively used to adapt dynamics from simulation to reality using a limited offline real dataset without rewards, and that \methodshort{} can be combined with visual adaptation methods to do so.

\section{Limitations and Future Work}

A potential limitation with \methodshort{} is that it excels at maintaining high target-environment performance over many updates because the residual avoids overfitting due to its low complexity. This suggests that only conceptually simple changes to dynamics may effectively be modeled with low amounts of data, warranting future investigation. 
We additionally want to explore if residual adaptation methods can be meaningfully applied to foundation world models, efficiently converting them from generators of plausible dynamics to generators of specific dynamics.

\section*{Acknowledgments}
Authors Lanier, Kim, and Karamzade were supported by a Hasso Plattner Foundation Fellowship.
This work was funded in part by the National Science Foundation (Award \#2321786).
We would like to thank Tony Liu and Zihao Zhu for their assistance in exploring early ideas related to this work.

\bibliography{paper}
\bibliographystyle{icml2025}

\newpage
\appendix
\onecolumn
\section{Actor-Critic Training and Deployment}
\label{sec:actor_critic_training}
\begin{figure*}[h]
   \centering
   \includegraphics[width=1.0\textwidth]{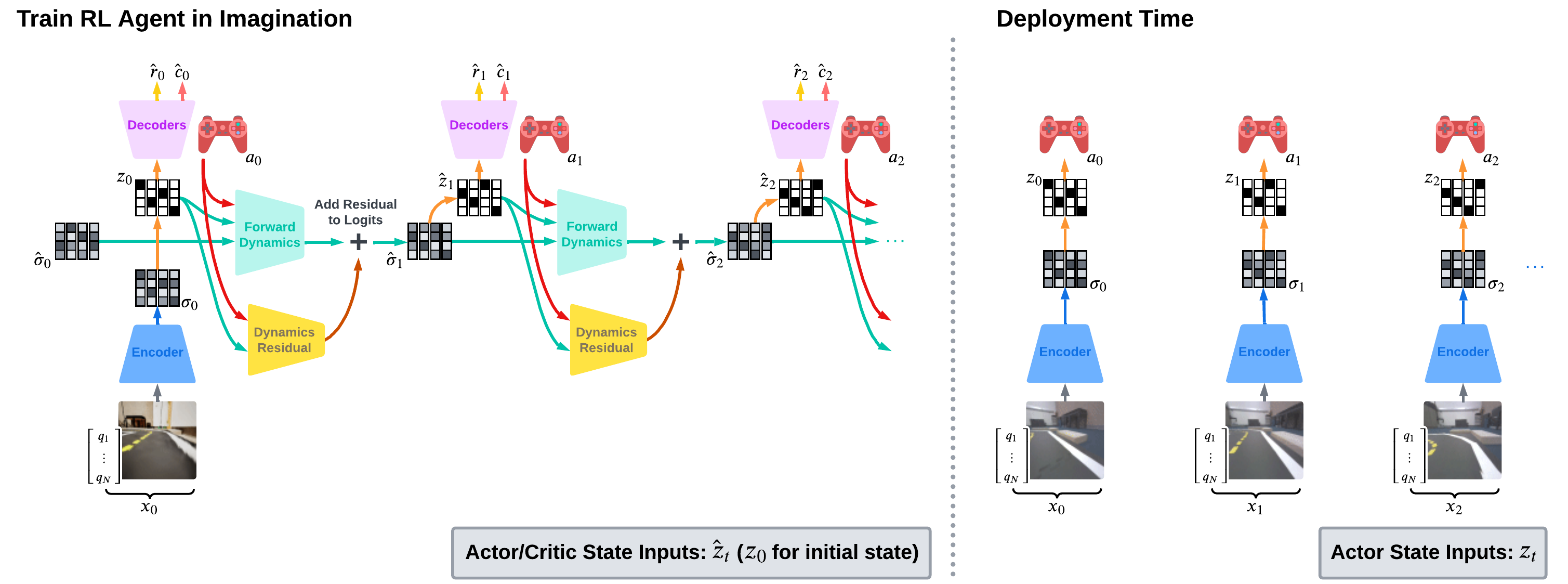}
   \caption{(\textbf{Left}) The actor and critic are trained by interacting with the world model. Starting from an environment state sampled from the replay buffer, the world model generates imagined rollouts using actions provided by the actor. The residual component is omitted during \archshort{} pretraining. (\textbf{Right}) At deployment, only the encoder and actor modules are utilized. The immediate environment state is processed by the encoder, and the actor generates an action based on $z_t$ sampled from $\sigma_t$.}
   \label{fig:actor_critic_training_and_deployment}
\end{figure*}

\section{Duckiebots Experiment Details}
\label{sec:duckiebot_details}

\paragraph{Simulation Reward Details}
In simulation, the agent is densely rewarded at each timestep with a value in $[0, 1]$ proportional to its projected velocity along the lane center unless its location is more than 5cm from the lane center, in which case it incurs a penalty of -1. When moving forward, we additionally provide a dense penalty proportional to egocentric yaw velocity to encourage turning while at speed. The simulation episode horizon is 200 steps, slightly more than enough time to complete a lap. We do not provide a termination signal when the horizon is reached. We terminate early with a done signal and a penalty of -100 if the agent drives off the track.

\begin{figure*}[h!]
   \centering
   \begin{subfigure}{0.48\textwidth}
       \centering
       \includegraphics[width=\textwidth]{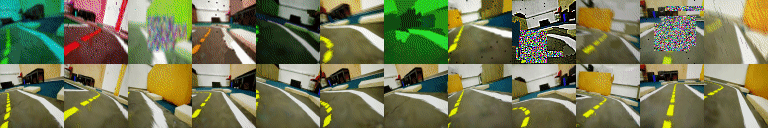}
       \caption{Simulation Images with Augmentations}
       \label{fig:sim_augmentations}
   \end{subfigure}
   \hfill %
   \begin{subfigure}{0.48\textwidth}
       \centering
       \includegraphics[width=\textwidth]{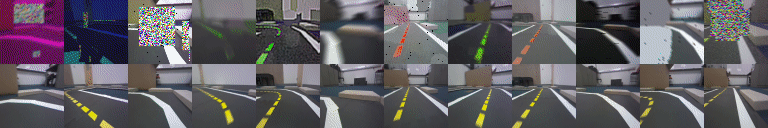}
       \caption{Real-world Images with Augmentations}
       \label{fig:real_augmentations}
   \end{subfigure}
   \caption{Comparison of image observations in simulation and the real world. (\textbf{Top}): Augmented images. (\textbf{Bottom}): Original images.}
   \label{fig:augmentations}
\end{figure*}

\paragraph{Image Augmentations} During simulation pretraining and offline adaptation to real data, we apply image augmentations to world model encoder inputs, but we still train decoder objectives on the original non-augmented images. Figure~\ref{fig:real_augmentations} shows original images (bottom) and their augmented counterparts (top) for both simulation and the real environment offline human demonstration dataset. In world model training for \archshort{}/\methodshort{} and Dreamer, we apply new image augmentations to each mini-batch after it is sampled from the experience buffer.

\newpage
\section{Learning Curves During Pretraining}
\label{sec:pretraining_curve}

\begin{figure*}[h!]
   \centering
   \includegraphics[width=1.0\textwidth]{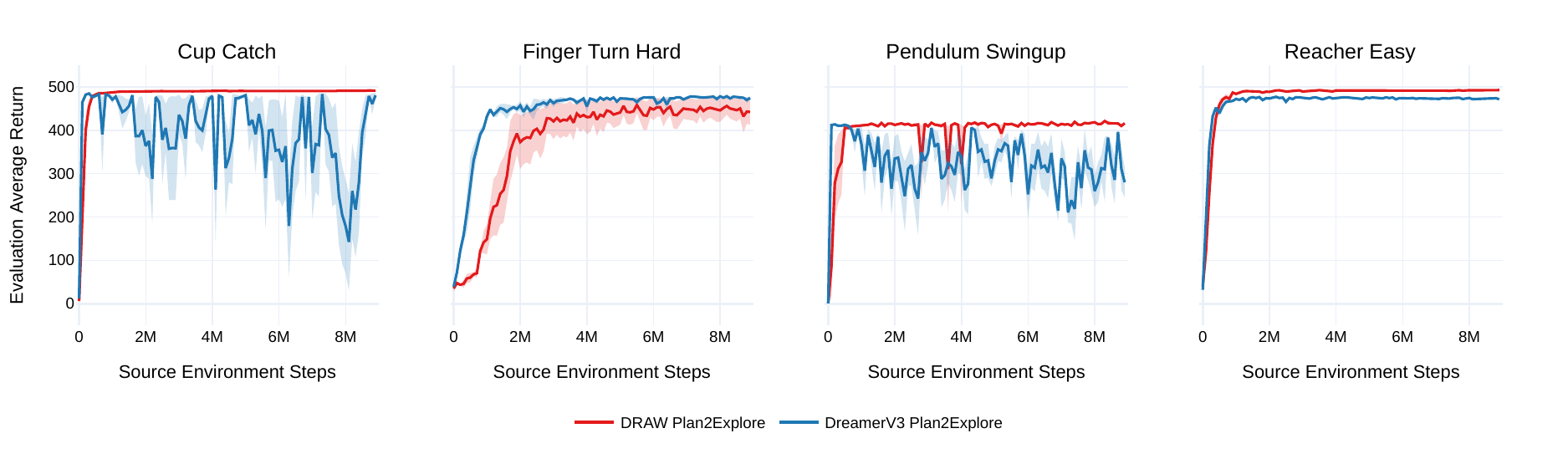}
   \caption{Training curves during pretraining for \archshort{} and DreamerV3 across four environments from DMC. Plan2Explore is used for data collection during pretraining. The mean and standard error are shown over 4 seeds.}
   \label{fig:main_mujoco_results_pretrain}
\end{figure*}

\begin{figure*}[h!]
   \centering
   \includegraphics[width=1.0\textwidth]{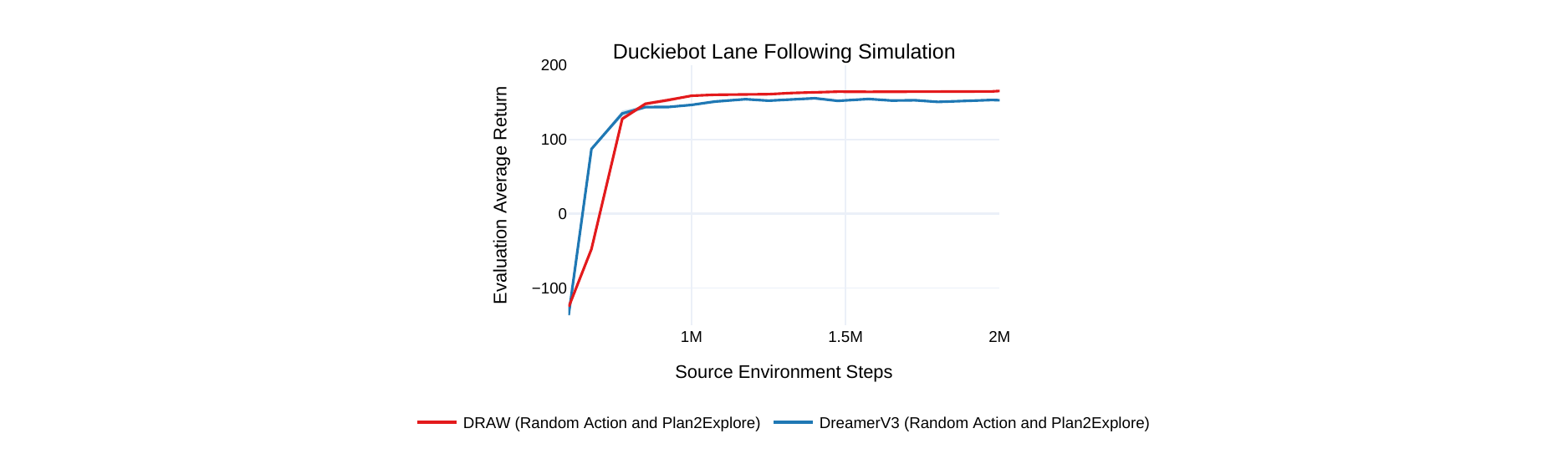}
   \caption{Training curves during pretraining for \archshort{} and DreamerV3 in the Duckiebot lane following simulation environment. Data collection is performed using random actions for the first 0.6M steps, followed by Plan2Explore for 1.4M steps. Each episode starts from a valid random position. The mean and standard error are shown over 4 seeds.}
   \label{fig:main_duckiebot_results_pretrain}
\end{figure*}

Figure~\ref{fig:main_mujoco_results_pretrain} and Figure~\ref{fig:main_duckiebot_results_pretrain} show training curves in the source environments in the DMC and Duckiebot domains, respectively. Both \archshort{} and DreamerV3 converge to similar performance in the source environments.

\section{Architectural Ablations}  
\label{sec:architecture_ablations}  

In this section, we examine how different choices for the inputs of the \archshort{} forward dynamics function $f_\theta$ and the \methodshort{} residual function $\delta_\psi$ affect transfer performance.  

\subsection{Forward Dynamics Inputs}  

\begin{figure*}[h!]
    \centering
    \begin{subfigure}[b]{1.0\textwidth}
        \centering
        \includegraphics[width=\textwidth]{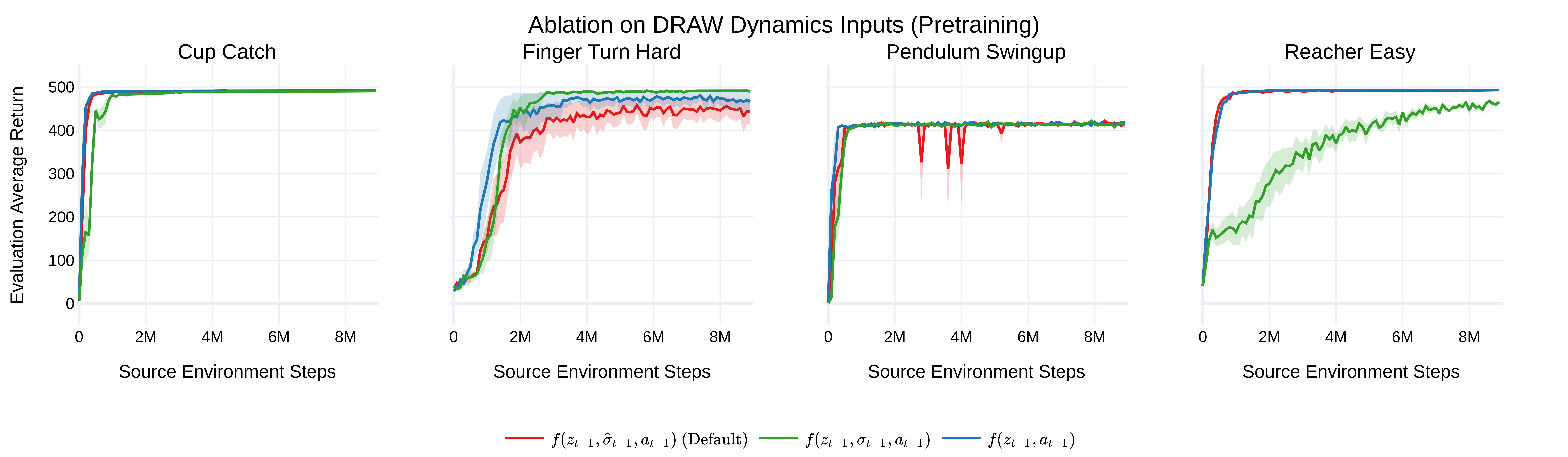}
        \caption{DMC source environment average return during \archshort{} pretraining with alternate dynamics function inputs.}
        \label{fig:architecture_ablation_pretrain}
    \end{subfigure}
    
    \vspace{0.5cm}  %

    \begin{subfigure}[b]{1.0\textwidth}
        \centering
        \includegraphics[width=\textwidth]{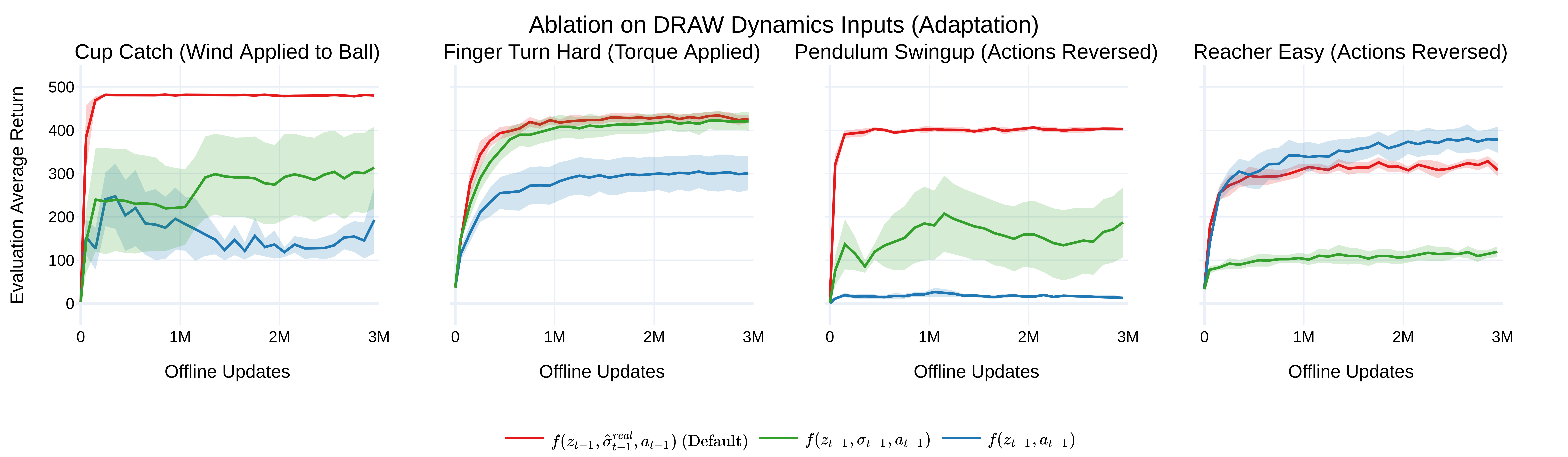} 
        \caption{DMC target environment average return during \methodshort{} residual adaptation with alternate dynamics function inputs.}
        \label{fig:architecture_ablation_transfer}
    \end{subfigure}
    
    \caption{Comparison of different dynamics function architectures of \archshort{} during pretraining (a) and adaptation (b).}
    \label{fig:architecture_ablation}
\end{figure*}

In the default \archshort{} architecture, $f_\theta$ is conditioned on the previous latent state $z_{t-1}$, the previous action $a_{t-1}$, and the additional input of the previous latent-state dynamic distribution $\hat{\sigma}_{t-1}$ (or $\hat{\sigma}_{t-1}^\textit{real}$ for \methodshort{}). In DMC environments, we compare this choice of inputs against two alternatives: (1) the minimal sufficient set $(z_{t-1}, a_{t-1})$, and (2) conditioning on the encoder latent-state distribution $\sigma_t = q_\theta(z_t|x_t)$. Figure~\ref{fig:architecture_ablation_pretrain} presents the performance of these different dynamics functions on source environments during \archshort{} Plan2Explore pretraining, while Figure~\ref{fig:architecture_ablation_transfer} shows their transfer performance on target environments during offline \methodshort{} adaptation.  

During pretraining, most input choices yield similar source-task performance. However, during adaptation, the default configuration, $f_\theta(z_{t-1}, \hat{\sigma}_{t-1}, a_{t-1})$, consistently outperforms the alternatives, achieving and maintaining higher performance in the target environments. We hypothesize that because including $\hat{\sigma}_{t-1}$ during world model pretraining facilitates gradient propagation over multiple timesteps, this inclusion enables the learning of features that improve long-term predictions.  

This advantage is achieved without increasing the residual's complexity, which could have otherwise negatively impacted transfer performance. During \methodshort{} adaptation, $\hat{\sigma}_{t-1}^\textit{real}$ serves as an input to $f_\theta(z_{t-1}, \hat{\sigma}_{t-1}^\textit{real}, a_{t-1})$. While conditioning on $\hat{\sigma}_{t-1}^\textit{real}$ increases the dimensionality of $f_\theta$'s input space, it has minimal impact on the complexity of the residual prediction $\delta_\psi$. Since $\hat{\sigma}_t^\textit{real}$ is already an output of the calibrated dynamics,  
\[
\hat{\sigma}_t^\textit{real} = \R{softmax}(f_\theta(z_{t-1}, \hat{\sigma}^\textit{real}_{t-1}, a_{t-1}) + \delta_\psi(z_{t-1}, a_{t-1})),
\]  
it can be included as an input to $f_\theta$ without increasing the dimensionality of the input or output spaces of $\delta_\psi$. This helps maintain the residual function’s simplicity, reducing the risk of overfitting.

\subsection{Residual Inputs}  

\begin{figure*}[h!]
   \centering
   \includegraphics[width=1.0\textwidth]{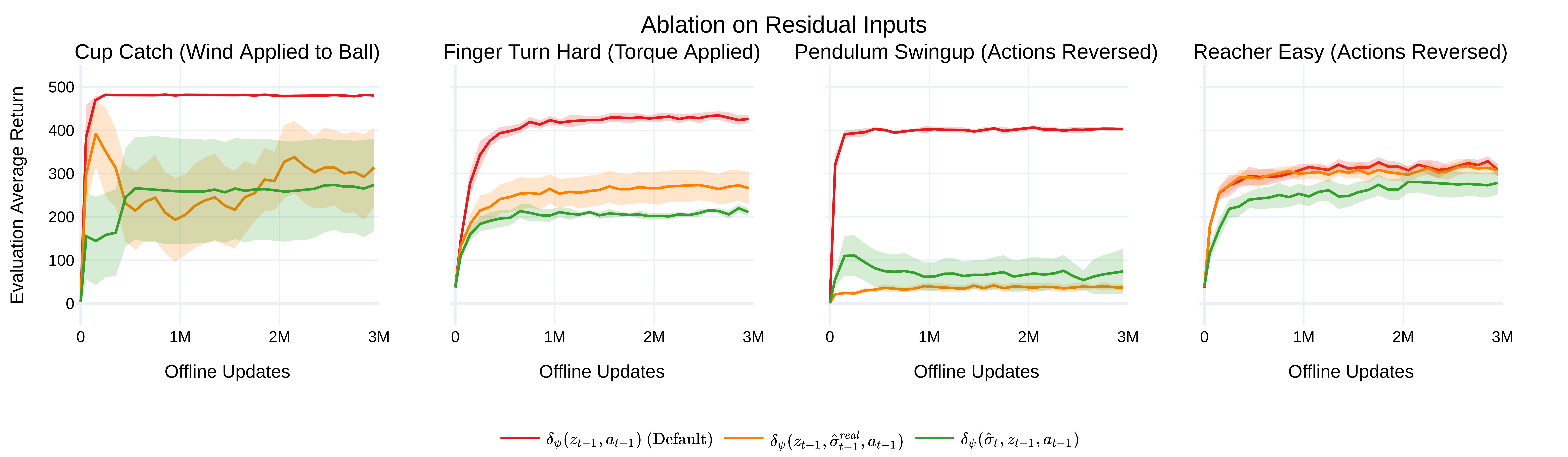}
   \caption{Comparison of different residual inputs for \methodshort.}
   \label{fig:res_ablation}
\end{figure*}

Next, in Figure~\ref{fig:res_ablation}, we compare the target environment transfer performance of our default residual function, $\delta_\psi(z_{t-1}, a_{t-1})$, against two alternative input configurations. The first, $\delta_\psi(z_{t-1}, \hat{\sigma}_{t-1}^\textit{real}, a_{t-1})$, conditions on the same inputs as $f_\theta$, while the second, $\delta_\psi(\hat{\sigma}_{t}, z_{t-1}, a_{t-1})$, additionally incorporates the original source environment dynamics predictions made by the frozen forward belief, $p_\theta(\hat{z}_t | z_{t-1}, \hat{\sigma}_{t-1}^\textit{real}, a_{t-1})$.

Although the additional inputs, $\hat{\sigma}_{t-1}^\textit{real}$ and $\hat{\sigma}_{t}$, could theoretically provide useful information for the residual prediction task, we observe that their inclusion leads to a decrease in target-environment performance. We hypothesize that conditioning the residual function on an added real-valued vector, alongside the discrete latent-state $z_{t-1}$, significantly expands the space of representable residual functions. Given the limited dataset, this increased complexity likely impairs generalization to the target domain.  

This result underscores the importance of bottlenecking state information through the compressed discrete representation $z_t$ for effective low-data adaptation.

\section{Latent Residual vs. New Dynamics Function}  
\label{sec:new_dyn_fn}  

In this section, we compare the \methodshort{} latent-state dynamics residual with an alternative adaptation method that also leverages frozen dynamics predictions learned from the source environment. Specifically, we contrast using a residual with learning a new replacement dynamics function, $g_\psi$, which optionally conditions on the outputs of the original source environment dynamics $f_\theta$. We evaluate three possible definitions for $g_\psi$:  

\begin{enumerate}
\item $\hat{\sigma}_t^\textit{real} = g_\psi(z_{t-1}, a_{t-1})$, where $g_\psi$ conditions on the same inputs as the \methodshort{} residual.
\item $\hat{\sigma}_t^\textit{real} = g_\psi(\hat{\sigma}_t, z_{t-1}, a_{t-1})$, where $g_\psi$ additionally conditions on the frozen \archshort{} predicted source dynamics distribution, $\hat{\sigma}_t = p_\theta(\hat{z}_t | z_{t-1}, \hat{\sigma}_{t-1}^\textit{real}, a_{t-1})$.
\item $\hat{\sigma}_t^\textit{real} = g_\psi(\hat{z}_{t}, z_{t-1}, a_{t-1})$, where $g_\psi$ additionally conditions on a discrete latent-state sample from the frozen \archshort{} source dynamics predictions, $\hat{z}_{t} \sim \R{MultiCategorical}(\hat{\sigma}_t)$ as in~\eqref{eq:forward_sample}.  
\end{enumerate}

To train the replacement dynamics function on the offline $M_\textit{real}$ dataset, we employ a dynamics loss term equivalent to~\eqref{eq:residual_loss} used by \methodshort{}:  

\begin{equation}  
    \label{eq:rp_dyn_lss}  
   \mathcal{L}_g(\psi) = \mathbb{E}_{q_{\bar{\theta}} (z_{1:T}|\zeta^\textit{real})}  
   \left[\sum_{t=1}^{T} \D[q_{\bar{\theta}}(z_t|x_t) || g_{\psi}(\hat{z}^\textit{real}_t|\bullet)]\right]
\end{equation}  

\begin{figure*}[h!]
   \centering
   \includegraphics[width=1.0\textwidth]{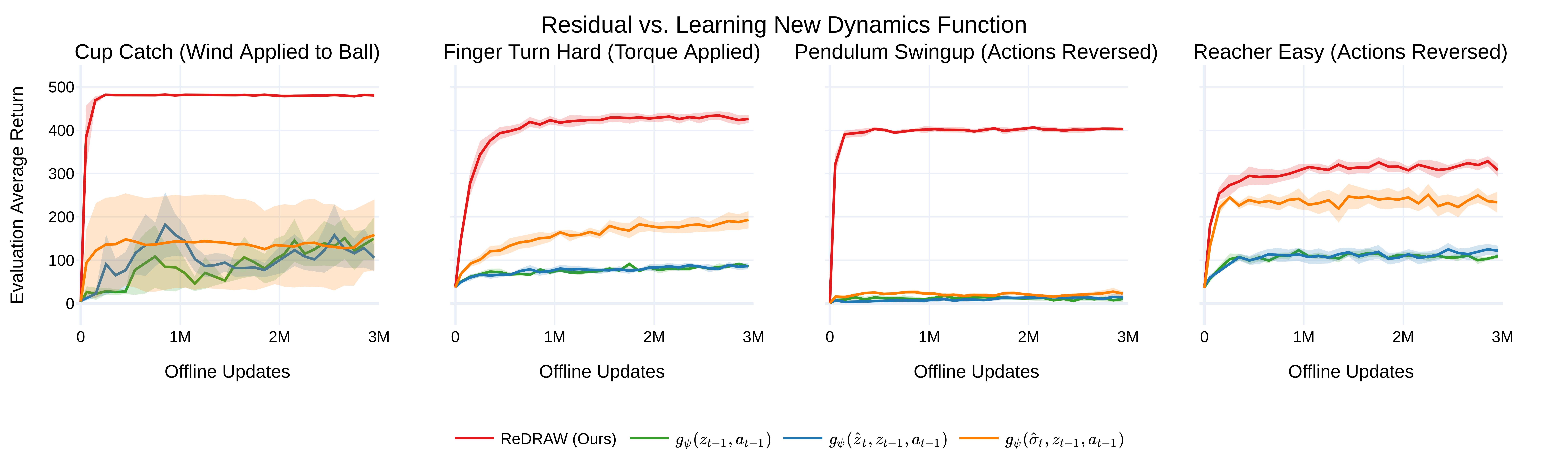}
   \caption{Comparison with a replacement dynamics function $g_\psi$ with the same small capacity as the residual network.}
   \label{fig:rp_dyn_ablation}
\end{figure*}

Figure~\ref{fig:rp_dyn_ablation} presents the average target environment return during adaptation for \methodshort{} and all considered replacement dynamics functions. The results show that \methodshort{} outperforms all variations of the replacement function baseline, including those that incorporate predictions from the frozen \archshort{} source dynamics function.  

From this experiment, we conclude that the residual operation, which modifies \archshort{} dynamics predictions without conditioning on them, is a key factor in achieving effective generalization to the target environment.

\section{DMC Experiment Details}
\label{sec:appendix:dmc_exp}

The state spaces for the DMC environments in this work consist of an image of the robot paired with a vector of ego-centric joint velocities. By providing a vector of velocities, we ensure full observability. We use an action repeat of two, meaning that each episode consists of 500 decision steps, equivalent to 1000 environment steps. Additionally, to preserve state-based rewards, we do not sum rewards over the environment steps skipped due to action repeat. 

Below, we describe each pair of source and target environments used in our DMC experiments. The source environment corresponds to the original DMC environment, while each target environment has modified dynamics:

\begin{itemize}
    \item \textbf{Cup Catch}: The agent controls a cup to catch a ball tethered by a string. In the target environment, a constant horizontal wind alters the ball's trajectory, requiring the agent to adapt by compensating for this external force.  
    \item \textbf{Finger Turn Hard}: The agent rotates a hinged spinner to a specified goal orientation. In the target environment, an external torque continuously drives the spinner, forcing the agent to counteract this disturbance to maintain control.  
    \item \textbf{Pendulum Swingup}: The agent swings a pendulum to an upright position. In the target domain, action effects are reversed, requiring the agent to invert its control policy.  
    \item \textbf{Reacher Easy}: The agent maneuvers a two-link arm to reach a target position. As in Pendulum Swingup, actions are inverted in the target environment, posing a challenge for direct policy transfer.  
\end{itemize}

\section{Hyperparameters}

We implement \archshort{} and \methodshort{} code as a modification to the official DreamerV3 implementation~\citep{dreamerv3_jax_code}. Except where otherwise stated, we use DreamerV3 default hyperparameters for all methods, including a batch size of 16, batch length of 64, and learning rates of \(1\times10^{-4}\) for the world model and \(3\times10^{-5}\) for the actor and critic. 
Additional parameters specific to our method or experiments are listed below.

\begin{table}[H]
    \centering
    \begin{tabular}{l l c}
        \toprule
        & \textbf{Hyperparameter} & \textbf{Value} \\
        
        \midrule
        \multirow{2}{*}{all methods} & pretraining replay buffer size  & 1e7  \\
             & online train ratio & 512 \\
             & Encoder/Decoder CNN Depth & 32 \\
             & Encoder/Decoder MLP hidden layers & 2 \\
             & MLP hidden units & 512 \\
             & image size & 64$\times$64$\times$3 \\
        \midrule
        \multirow{6}{*}{\archshort{}/\methodshort{}}& $K$ (number of categorical distributions) & 256 \\
        & $N$ (number of categorical classes) & 4 \\
        & imagination horizon for actor-critic training & 40 \\
        & $\beta_{pred}$ & 1.0 \\
        & $\beta_{dyn}$ & 1.5 \\
        & $\beta_{rep}$ & 0.5 \\
        & residual learning rate & 1e-2 \\
        & forward dynamics MLP hidden layers & 1 \\
        & residual MLP hidden layers & 1 \\
        & residual MLP hidden units & 256 \\

        \bottomrule
    \end{tabular}
    \caption{Modified or newly introduced hyperparameters used in experiments.}
    \label{tab:hyperparameters}
\end{table}

\end{document}